\documentclass[pdflatex,sn-mathphys]{sn-jnl}% Math and Physical Sciences Reference Style
\jyear{2021}%

\theoremstyle{thmstyleone}%
\theoremstyle{thmstyletwo}%

\theoremstyle{thmstylethree}%

\usepackage{graphicx}
\usepackage{xcolor}
\usepackage{hyperref}

\usepackage{pifont}
\usepackage{xcolor}
\newcommand{\change}[1]{\textcolor{black}{#1}}
\newcommand{\changetwo}[1]{\textcolor{black}{#1}}

\usepackage{multirow}

\begin{document}

\title[Paced-Curriculum Distillation with Prediction and Label Uncertainty]{Paced-Curriculum Distillation with Prediction and Label Uncertainty for Image Segmentation}

\author{Mobarakol Islam$^{1\dagger}$, Lalithkumar Seenivasan$^{2\dagger}$, S P Sharan$^{3}$, V.K. Viekash$^{4}$, Bhavesh Gupta$^{5}$, Ben Glocker$^{1}$ and Hongliang Ren$^{2,6}$\footnote{ Corresponding to Hongliang Ren hlren@ieee.org.\\
    $^{\dagger}$ Equal contribution\\  
    $^{1}$Dept. of Computing, Imperial College London, United Kingdom. \\
    $^{2}$Dept. of Biomedical Engineering, National University of Singapore, Singapore.\\
    $^{3}$Dept. of Electronics and Communication Engineering, National Institute of Technology Tiruchirappalli, India.\\
    $^{4}$Dept. of Instrumentation and Control Engineering, National Institute of Technology Tiruchirappalli, India.\\
    $^{5}$Dept. of Mechanical Engineering, Indian Institute of Technology, Delhi, India\\
    $^{6}$Dept. of Electrical Engineering and Shun Hing Institute of Advanced Engineering, The Chinese University of Hong Kong.}

}

\abstract{
\textbf{Purpose:}
In curriculum learning, the idea is to train on easier samples first and gradually increase the difficulty, while in self-paced learning, a pacing function defines the speed to adapt the training progress. While both methods heavily rely on the ability to score the difficulty of data samples, an optimal scoring function is still under exploration.

\textbf{Methodology:} 
Distillation is a knowledge transfer approach where a teacher network guides a student network by feeding a sequence of random samples. We argue that guiding student networks with an efficient curriculum strategy can improve model generalization and robustness. For this purpose, we design an uncertainty-based paced curriculum learning in self-distillation for medical image segmentation. We fuse the prediction uncertainty and annotation boundary uncertainty to develop a novel paced-curriculum distillation (P-CD). We utilize the teacher model to obtain prediction uncertainty and spatially varying label smoothing with Gaussian kernel to generate segmentation boundary uncertainty from the annotation. We also investigate the robustness of our method by applying various types and severity of image perturbation and corruption. 

\textbf{Results:} 
The proposed technique is validated on two medical datasets of breast ultrasound image segmentation and robot-assisted surgical scene segmentation and achieved significantly better performance in terms of segmentation and robustness.

\textbf{Conclusion:}
P-CD improves the performance and obtains better generalization and robustness over the dataset shift. While curriculum learning requires extensive tuning of hyper-parameters for pacing function, the level of performance improvement suppresses this limitation. 
}


\maketitle

\section{Introduction}\label{Introduction}
Inspired by the cognitive process of humans and animals, numerous curriculum learning (CL)~\cite{bengio2009curriculum,sinha2020curriculum} and self-paced learning (SPL)~\cite{kumar2010self,jiang2015self} techniques are introduced in machine learning. While both the CL and SPL techniques heavily rely on difficulty measurement techniques and ranking among the samples, an ideal ranking method is rarely available. Several methods attempt to design difficulty measurement techniques using confidence score~\cite{zhao2021knowledge}, multi-raters agreement~\cite{Wei_2021_WACV}, and feature smoothing~\cite{sinha2020curriculum} in classification tasks. However, most of these works focus on classification tasks where it is easy to drop the harder images (samples). In segmentation or pixel-wise classification, it is more complex to build a curriculum strategy where spatial dependency is crucial among the local structures. Specifically in medical imaging, object boundary label is always ambiguous, even by human experts. In this work, we design a novel curriculum strategy using prediction and annotation boundary uncertainty in self-distillation for medical image segmentation.

Most recently, CL and SPL techniques have proven superior with distillation in vision and language classification tasks. In terms of scoring techniques, the confidence score uses to rank the input examples~\cite{zhao2021knowledge}, the mirror decent technique uses to obtain sequential optimization~\cite{shi2021follow}, fixed lineups to increase the difficulty~\cite{ye2020towards}, Sentence length and coherence between dialog pairs use to determine sample complexity~\cite{zhu2021combining} and less imbalance subset uses for self-paced expert selection and self-paced instance selection~\cite{xiang2020learning} to build curriculum or self-paced knowledge distillation. However, most of these works focus on classification tasks where measuring sample difficulty without considering spatial correlation and local structure should not be compatible with semantic segmentation. Unlike distillation, both uncertainty and confidence-based sample ranking techniques are found to be the most common scoring method to design curriculum learning due to simplicity. For example, in the applications of classification~\cite{wang2019dynamic,chang2017active}, object localization~\cite{kumar2010self}, neural machine translation~\cite{zhou2020uncertainty,zhang2018empirical} and reinforcement learning~\cite{zhang2020automatic}. Nonetheless, the effectiveness of pixel-wise uncertainty-based curriculum strategy is less clear in semantic segmentation. The underlying spatial dependencies among the semantic object limit the correctness of pixel-wise difficulty measurement using predicted uncertainty independently. In addition, there is obvious uncertainty about anatomical structure in medical imaging which can be considered as difficult pixels regardless of \change{predicted uncertainty (PU)~\cite{lee2020structure, islam2021spatially, tang2022unified}.} 

In this work, we design paced-curriculum distillation (P-CD) by fusing \change{ prediction uncertainty (PU) and annotation boundary uncertainty (BU)} to determine pixel-wise difficulty scores for self-paced curriculum strategy. We use a teacher network to obtain the prediction uncertainty and spatially varying label smoothing~\cite{islam2021spatially} with Gaussian kernel to generate segmentation boundary uncertainty from the annotation. In P-CD, curriculum strategy allows guiding the student network to learn the easier pixels at the beginning of training and gradually increases the difficulty level and self-paced learning controls the speed of student network learning by feeding a certain ratio of samples with epochs. We validate the proposed method with two medical image segmentation datasets of robot-assisted surgical scene segmentation (Robotic Surgery)~\cite{allan20202018} and breast ultrasound tumor segmentation (BUS)~\cite{al2020dataset}. We conduct extensive robustness experiments by corrupting the datasets with various perturbation techniques in five different severity levels. Our contributions can be summarized as follows:
\begin{itemize}
    \item[--] Introduce a novel paced-curriculum distillation (P-CD) that fuses prediction uncertainty and annotation boundary uncertainty to determine pixel-wise
difficulty score, including structural dependency.
    \item[--] P-CD validates on two medical image segmentation datasets and the results suggest the superiority of the model over the baselines.
    \item[--] P-CD also demonstrates robust performance on the progressive perturbations with different severity levels.
\end{itemize}

\section{Proposed Method} \label{methodology}

\begin{figure*}[!hb]
    \centering
    \includegraphics[width=.95\linewidth]{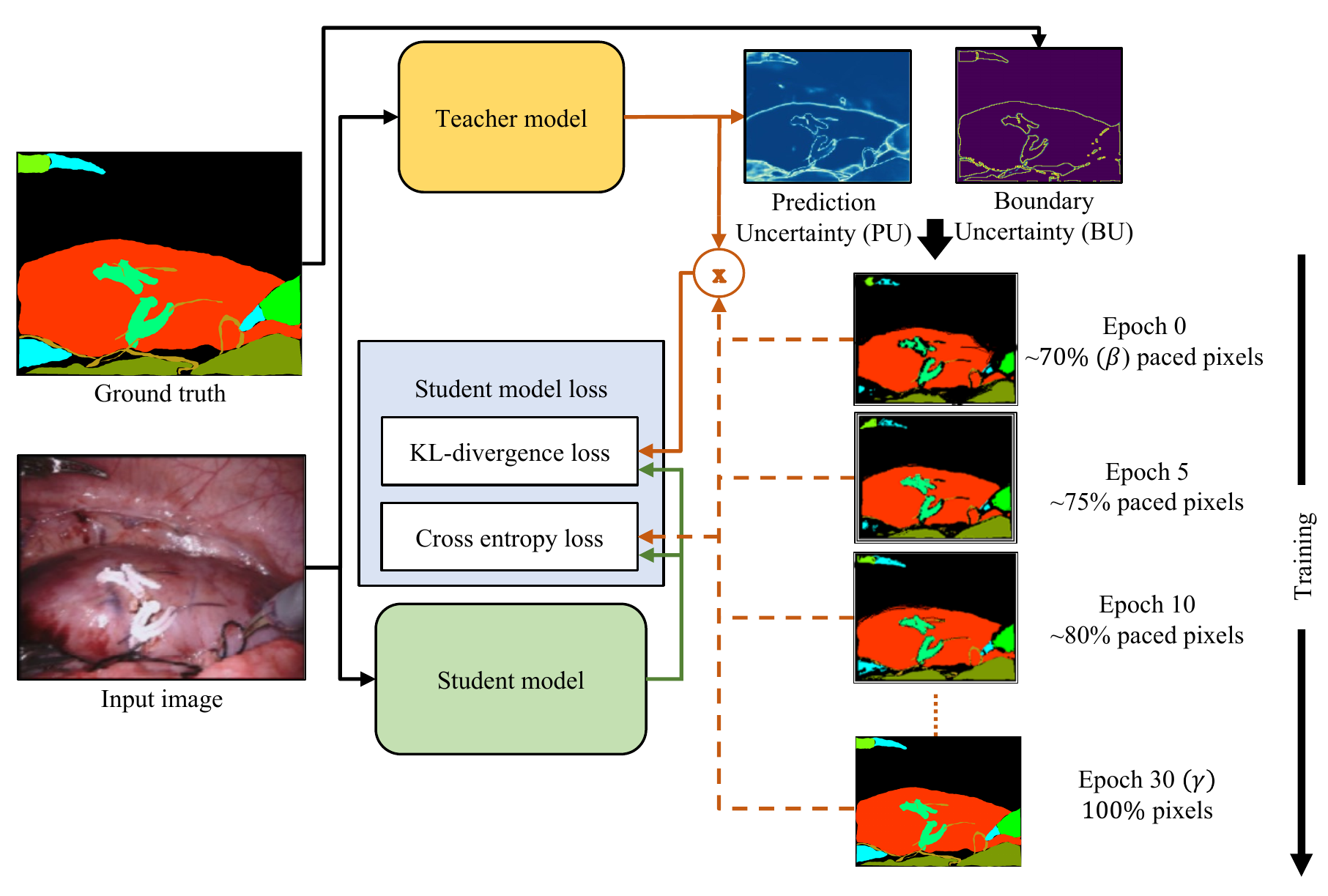}
    \caption{Paced-curriculum distillation (P-CD), an uncertainty-based paced-curriculum learning in self-distillation for segmentation. Given the input image and label, the input is first propagated through the teacher model. Then \change{calculate the pixel-wise difficulty score with the fusion of the predicted uncertainty from the teacher and label boundary uncertainty. Overall, curriculum learning (CL) guides the student to learn easier to harder pixels during training and self-paced learning controls the speed of the student network learning by feeding a certain ratio of pixels with epochs.}}
    \label{fig:pcd}
\end{figure*}

\subsection{Preliminaries}

\subsubsection{Distillation}
In distillation, besides training the student model against the true label, it is also trained to mimic the teacher model’s prediction probabilities ~\cite{hinton2015distilling}. This is achieved by training the model on both the cross-entropy loss ($loss_{CE}$) against the true label and the KL divergence loss ($loss_{KL}$) between the student and teacher logits \cite{kim2021self}. Training on $loss_{KL}$ enforces the student model to mimic similar feature representations to the teacher model. The self-distillation (SD) loss ($loss_{SD}$) is theorized as: $loss_{SD} = loss_{CE} + loss_{KL}$. If the image size is [$H$x$W$] then $loss_{SD}$ will be a matrix of the same size ($loss_{SD} \in \mathbb{R}^{H \times W}$) for pixel-wise loss before averaging in segmentation task. 

\subsubsection{Uncertainty}
Prediction uncertainty can interpret the confidence level of a model to recognize a sample. Alternatively, an easier sample produces lower uncertainty or higher confidence during the prediction. Therefore, prediction uncertainty (PU) can be easily determined as: 
\change{
\begin{equation}
     PU = (1 - confidence)
    \label{eq:pu}
\end{equation}
}
To generate boundary uncertainty (BU) from label, spatially varying label smoothing (SVLS)~\cite{islam2021spatially} design a weight matrix, with the \change{ $3 \times 3$ Gaussian kernel of the standard deviation, $\sigma = 1$ and modify the kernel to make sure center is equal to the sum of the neighbouring points and convolves across the OneHot label (OHLabel). } 
\change{
\begin{equation}
     BU = 1 - SVLS^{weight}(\sigma) \circledast OHLabel
    \label{eq:bu}
\end{equation}
}

In this work, we utilize the SVLS~\cite{islam2021spatially} to obtain BU and fuse it with PU to design a novel curriculum scheme in the distillation process.

\subsection{Paced-Curriculum Distillation (P-CD)} \label{P-CD}
Curriculum learning, in most cases, is implemented on classification tasks as it is easier to discard tough images (samples). However, choosing which images to drop gets complicated for segmentation tasks as some pixels could be easier to learn than others in every image. One way to address this is to find the average difficulty of the pixels in an image that could be used as a parameter to quantify the difficulty of an image. However, this leads to a new problem as most images consist of pixels of varying difficulty. Eventually, this would still result in model learning both easy and hard pixels simultaneously, resulting in sub-optimal model convergence. To address this, we reformulate segmentation tasks as pixel-wise classification with images depicting mini-batches of pixels. 

We design Paced-Curriculum Distillation (P-CD) by fusing prediction uncertainty from the teacher networks and boundary uncertainty from labels to determine pixel-wise difficulty for curriculum strategy. During distillation, we control the loss calculation by allowing easier pixels to learn at the beginning of the training and gradually including harder pixels over the training, as shown in Fig~\ref{fig:pcd}. We also design a pacing function to guide the suitable learning speed by determining the pixel ratio to calculate loss over training epochs. Overall procedures can be formulated as below.

\subsubsection{Confidence Calibration}
As confidence score is one of the major criteria in our curriculum design, we calibrate the confidence score on the teacher model to obtain accurate uncertainty.  We apply, temperature scaling~\cite{guo2017calibration}, a commonly used calibration technique to calibrate the model. In temperature scaling, an optimal temperature $T$ is used to scale the logits $z$  as $(z/T)$ to obtain calibrated confidence.

\subsubsection{Curriculum Objective Function}
The effect of our curriculum directly influences the objective function of the student network during distillation. We drop the pixels with low confidence or high uncertainty in loss calculation at the initial stage of the training. For this purpose, we determine pixel-wise uncertainty weight matrices for both prediction uncertainty ($W_{PU}$) from calibrated teacher model (as equation~\ref{eq:pu}) and boundary uncertainty ($W_{BU}$) from annotation label (as equation~\ref{eq:bu}) by applying a threshold \change{$\mu$. This initial value of the threshold and increment criteria depend on the initial ratio of the pixels and end epoch of the curriculum learning (more details are in section~\ref{pacing})}. If $\mu$ is the uncertainty threshold (which determine by the pacing function), then binary weights matrices $W_{PU}$ and $W_{BU}$ can be calculated as-

\begin{equation}
    W_{PU} \text{ or } W_{BU}  =
    \begin{cases}
        0 ,& \text{if }W_{PU} \text{ or } W_{BU} >= \mu \\
        1 ,& \text{otherwise}
    \end{cases}
\label{eq:pcd_pu_bu}
\end{equation}

If the pixel-wise vanilla distillation loss matrix $loss_{SD} \in \mathbb{R}^{H \times W}$, prediction and boundary uncertainty weight matrices $W_{PU} \in \mathbb{R}^{H \times W}$ and $W_{BU} \in \mathbb{R}^{H \times W}$, then our paced-curriculum distillation loss matrix (before averaging) will be-
\begin{equation}
    loss_{PCD} = W_{BU} \odot W_{PU} \odot loss_{KD}
    \label{eq:pcd}
\end{equation}

\subsubsection{Pacing Function}
\label{pacing}
We design the pacing function based on two important criteria such as the initial ratio of data, $\beta$, and the end epoch of curriculum training, $\gamma$. Based on $\beta$, we can determine the initial uncertainty threshold \change{$\mu_{init}$} from the teacher network. Then, threshold update $\mu_{update}$ can be obtained from the following equation-
\change{
\begin{flalign}
\label{eq:pacing}
\mu_{update} &= \frac{1-\mu_{init}}{\gamma / E_{interval}}\\
\mu &= \mu_{init} + \mu_{update}\\
\mu &= min(\mu, 1)
\end{flalign}
}
Where $E_{interval}$ is the epoch interval, we want to update the uncertainty to include harder pixels over the training period. We use $E_{interval} = 5$ by following~\cite{sinha2020curriculum}. \change{When the uncertainty threshold, $\mu = 1$, then the student model will consider all the pixels to calculate loss during training.} 

\section{Experiments} \label{Experiments}

\subsection{Datasets}

\subsubsection{Robotic Surgery}
The MICCAI18 instrument segmentation challenge~\cite{allan20202018} dataset consists of $15$ video sequences for training and $4$ video sequences for testing. Each sequence consists of $149$ frames which are resized to $224$ x $224$ in this work. For pixel-wise segmentation, the dataset consists of $12$ classes of Kidney Parenchyma, Covered Kidney, Instrument Shaft, Instrument Clasper, Instrument Wrist, Thread, Suturing Needle, US Probe, Small Intestine, Suction Instrument, Clamps, and Background.

\subsubsection{BUS}
Breast ultrasound tumor segmentation (BUS)~\cite{al2020dataset} is a publicly available dataset\footnote{https://www.kaggle.com/datasets/aryashah2k/breast-ultrasound-images-dataset} that consists of 780 breast ultrasound images and corresponding tumor labels. All these images and labels are resized to 224 × 224 and split to 80/20 ratio for the training and validation purposes of this work. 

\subsection{Implementation Details}
We first train some of the independent and identically distributed (IID) models like vanilla UNet~\cite{ronneberger2015unet}, LinkNet~\cite{chaurasia2017linknet}, DeepLabv3+~\cite{chen2018encoder} as teacher models for rest of the experiment. We then train the SD and proposed the P-CD method. As accurate uncertainty or confidence is crucial to determine pixel difficulty, we calibrate each teacher model using the commonly used technique temperature scaling~\cite{guo2017calibration}. Overall, the models are trained on Adam optimizer with a batch size of 16, and the number of epochs is taken to be 150.

\section{Results} \label{results}

\begin{table}[!ht]
\centering
\caption{Performance comparison of proposed P-CD over the model train on vanilla SD and i.i.d. for three state-of-the-art architectures of UNet, DeepLabV3+, and LinkNet.}
\label{tab:main_table}
\scalebox{0.925}{
\begin{tabular}{c|c|ccc|ccc}
\toprule
\multirow{2}{*}{Methods}                               &           & \multicolumn{3}{c|}{\textbf{Robotic surgery}}                                                  & \multicolumn{3}{c}{\textbf{BUS}}                                                              \\ \cline{2-8} 
                                                       & Metrics   & \multicolumn{1}{c|}{\textbf{i.i.d.}} & \multicolumn{1}{c|}{\textbf{SD}}     & \textbf{P-CD}    & \multicolumn{1}{c|}{\textbf{i.i.d.}} & \multicolumn{1}{c|}{\textbf{SD}}     & \textbf{P-CD}   \\
\midrule
\multirow{3}{*}{UNet~\cite{ronneberger2015unet}}       & DSC       & \multicolumn{1}{c|}{0.6251}          & \multicolumn{1}{c|}{0.6445}          & \textbf{0.6644}  & \multicolumn{1}{c|}{0.7620}          & \multicolumn{1}{c|}{0.7676}          & \textbf{0.7897} \\ \cline{2-8} 
                                                       & IoU       & \multicolumn{1}{c|}{0.5827}          & \multicolumn{1}{c|}{0.6010}          & \textbf{0.6237}  & \multicolumn{1}{c|}{0.6921}          & \multicolumn{1}{c|}{0.6919}          & \textbf{0.7191} \\ \cline{2-8} 
                                                       & Precision & \multicolumn{1}{c|}{0.6890}          & \multicolumn{1}{c|}{0.7212}          & \textbf{0.7535}  & \multicolumn{1}{c|}{0.8696}          & \multicolumn{1}{c|}{\textbf{0.9092}} & 0.9042          \\
\midrule
\multirow{3}{*}{DeepLabV3+~\cite{chen2018encoder}}     & DSC       & \multicolumn{1}{c|}{0.6656}          & \multicolumn{1}{c|}{0.6858}          & \textbf{0.6929}  & \multicolumn{1}{c|}{0.7382}          & \multicolumn{1}{c|}{0.7611}          & \textbf{0.7722}               \\ \cline{2-8} 
                                                       & IoU       & \multicolumn{1}{c|}{0.6237}          & \multicolumn{1}{c|}{0.6437}          & \textbf{0.6503}  & \multicolumn{1}{c|}{0.6556}          & \multicolumn{1}{c|}{0.6762}          & \textbf{0.6860}               \\ \cline{2-8} 
                                                       & Precision & \multicolumn{1}{c|}{0.7511}          & \multicolumn{1}{c|}{\textbf{0.7829}} & 0.7803           & \multicolumn{1}{c|}{0.8389}          & \multicolumn{1}{c|}{\textbf{0.8557}}          & 0.8328               \\ 
\midrule
\multirow{3}{*}{LinkNet~\cite{chaurasia2017linknet}}   & DSC       & \multicolumn{1}{c|}{0.6442}          & \multicolumn{1}{c|}{0.6614}          & \textbf{0.6715}  & \multicolumn{1}{c|}{0.7261}          & \multicolumn{1}{c|}{0.7753}          & \textbf{0.7777} \\ \cline{2-8} 
                                                       & IoU       & \multicolumn{1}{c|}{0.6071}          & \multicolumn{1}{c|}{0.6236}          & \textbf{0.6332}  & \multicolumn{1}{c|}{0.6547}          & \multicolumn{1}{c|}{0.6962}          & \textbf{0.7045} \\ \cline{2-8} 
                                                       & Precision & \multicolumn{1}{c|}{0.7486}          & \multicolumn{1}{c|}{0.7419}          & \textbf{0.7707}  & \multicolumn{1}{c|}{0.8448}          & \multicolumn{1}{c|}{0.6962}          & \textbf{0.8703} \\ 
\bottomrule
\end{tabular}}
\end{table}

\begin{table}[!t]
\centering
\caption{\change{Significant test (p-value) on DSC and IoU improvement when trained using (i) i.i.d vs SD, (ii) P-CD vs SD, (iii) and P-CD vs i.i.d, on both the BUS and Robotic surgery datasets.}}
\label{tab:pvalue}
\scalebox{0.925}{
\change{
\begin{tabular}{ccccccc}
\toprule
\multicolumn{1}{c|}{\multirow{3}{*}{\textbf{Dataset}}} & \multicolumn{3}{c|}{\textbf{DSC}}                                                                                                                                                                                                                                                                                      & \multicolumn{3}{c}{\textbf{IoU}}                                                                                                                                                                                                                                                                  \\ \cline{2-7} 
\multicolumn{1}{c|}{}                                  & \multicolumn{1}{c|}{\multirow{2}{*}{\textbf{\begin{tabular}[c]{@{}c@{}}i.i.d \\ vs SD\end{tabular}}}} & \multicolumn{1}{c|}{\multirow{2}{*}{\textbf{\begin{tabular}[c]{@{}c@{}}SD \\ vs P-CD\end{tabular}}}} & \multicolumn{1}{c|}{\multirow{2}{*}{\textbf{\begin{tabular}[c]{@{}c@{}}i.i.d \\ vs P-CD\end{tabular}}}} & \multicolumn{1}{c|}{\multirow{2}{*}{\textbf{\begin{tabular}[c]{@{}c@{}}i.i.d \\ vs SD\end{tabular}}}} & \multicolumn{1}{c|}{\multirow{2}{*}{\textbf{\begin{tabular}[c]{@{}c@{}}SD \\ vs P-CD\end{tabular}}}} & \multirow{2}{*}{\textbf{\begin{tabular}[c]{@{}c@{}}i.i.d \\ vs P-CD\end{tabular}}} \\
\multicolumn{1}{c|}{}                                  & \multicolumn{1}{c|}{}                                                                                 & \multicolumn{1}{c|}{}                                                                                & \multicolumn{1}{c|}{}                                                                                   & \multicolumn{1}{c|}{}                                                                                 & \multicolumn{1}{c|}{}                                                                                &                                                                                    \\ \midrule
\multicolumn{1}{c|}{\textbf{BUS}}                      & \multicolumn{1}{c|}{\textbf{0.042}}                                                                           & \multicolumn{1}{c|}{\textbf{0.073}}                                                                        & \multicolumn{1}{c|}{\textbf{0.016}}                                                                             & \multicolumn{1}{c|}{0.104}                                                                          & \multicolumn{1}{c|}{0.127}                                                                        & \textbf{0.042}                                                                            \\ \midrule
\multicolumn{1}{c|}{\textbf{Robotic surgery}}          & \multicolumn{1}{c|}{0.161}                                                                           & \multicolumn{1}{c|}{0.224}                                                                          & \multicolumn{1}{c|}{\textbf{0.048}}                                                                            & \multicolumn{1}{c|}{0.173}                                                                           & \multicolumn{1}{c|}{0.212}                                                                          & \textbf{0.046}                                                                            \\ \bottomrule
\end{tabular}}}
\end{table}

\begin{figure}[!t]
    \centering
    \includegraphics[width=0.95\linewidth]{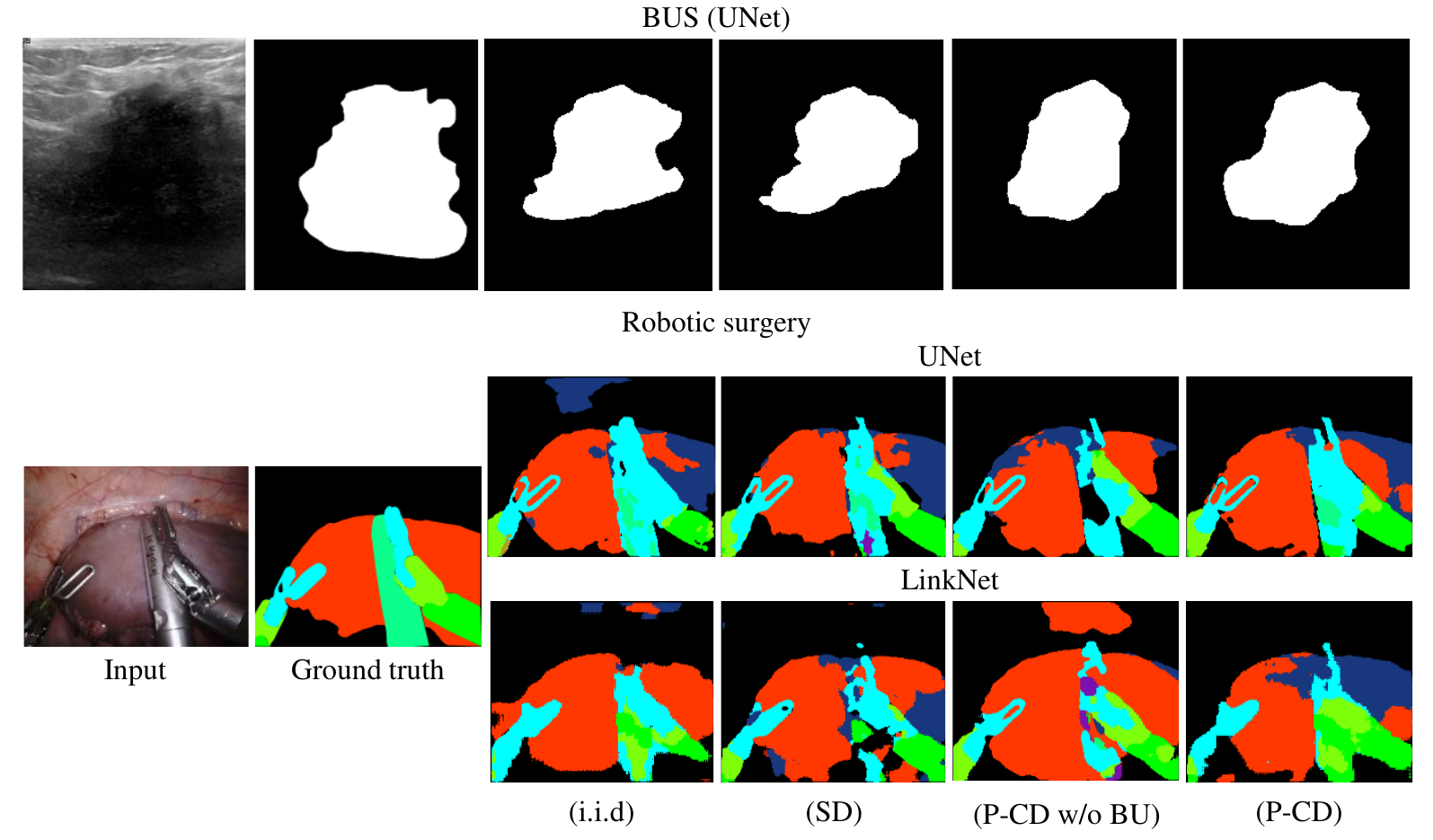}
    \caption{Qualitative visualization of our P-CD or P-CD w/o BU against the ground truth (GT), teacher model, and student model trained using self-distillation (SD). \change{The colors red, dark blue, light blue, and green mean kidney-parenchyma, covered-kidney, instrument-clasper, and wrist, respectively, in the robotic surgery dataset. In the BUS dataset, breast tumors are annotated as white color.}}
    \label{fig:qualitative_analysis}
\end{figure}

The proposed method, P-CD, has been extensively validated on three state-of-the-art architectures of UNet~\cite{ronneberger2015unet}, DeepLabV3+~\cite{chen2018encoder}, and LinkNet~\cite{chaurasia2017linknet} and two medical image segmentation datasets and three commonly used evaluation metrics of e dice similarity coefficient (DSC),  mean intersection over union (IoU), and precision. The results are presented in the Table~\ref{tab:main_table} and Fig.~\ref{fig:qualitative_analysis}.

\begin{figure}[!t]
    \centering
    \includegraphics[width=0.95\linewidth]{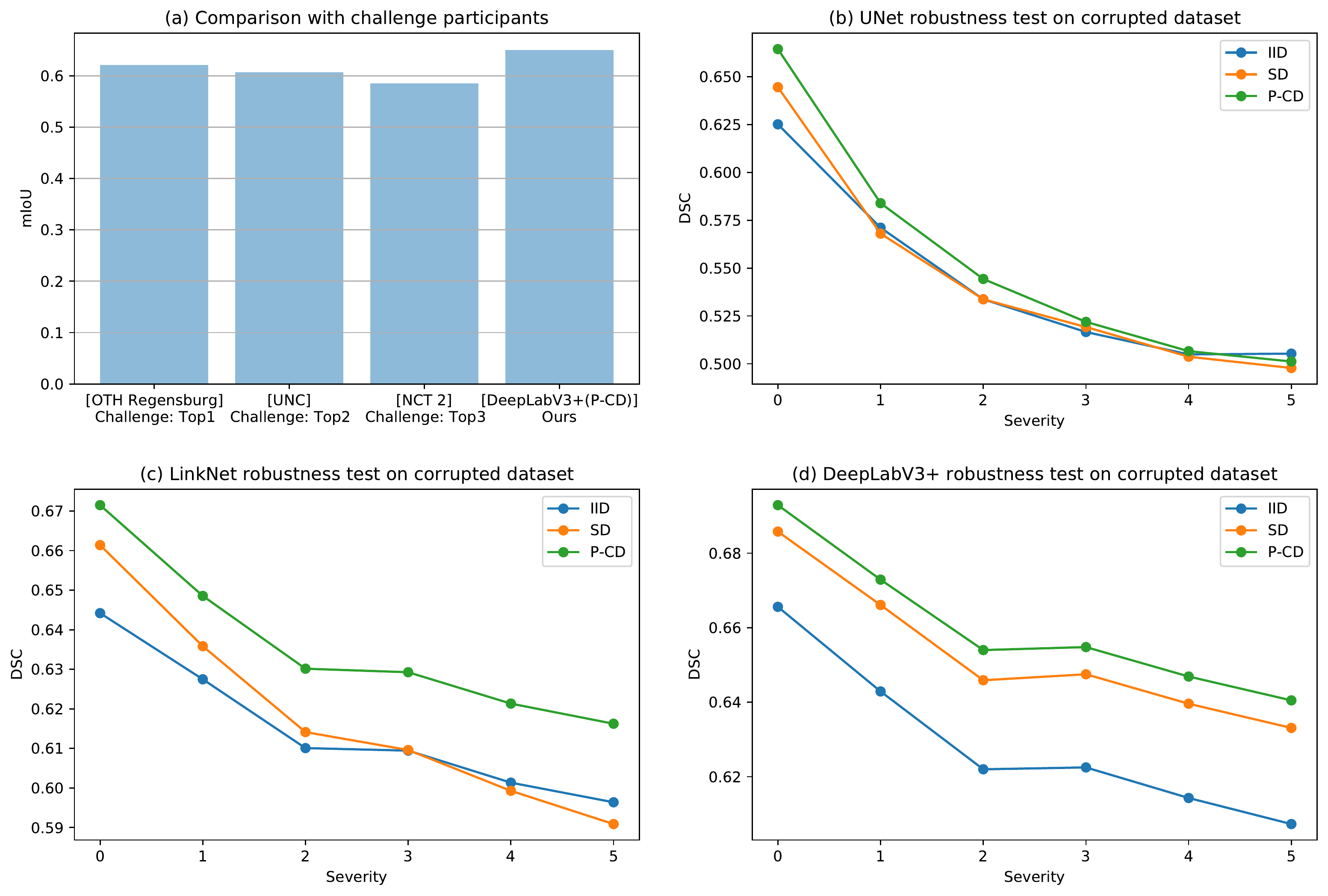}
    \caption{\change{Comparison and robustness studies: (a) Comparison with the challenge, (b) Robustness test on the Robotic surgery dataset with UNet model, (c) Robustness test on the Robotic surgery dataset with LinkNet model, (d) Robustness test on the Robotic surgery dataset with DeepLabv3+ model.}}
    \label{fig:ComparisionandRobustness}
\end{figure}

From Table. \ref{tab:main_table}, it is observed that the student model trained on the proposed P-CD outperforms both \change{independent and identically distributed (i.i.d.)} training model and vanilla SD version. Our method obtains DSC improvement of around 2-4\% over i.i.d. for all the architectures on both datasets. \changetwo{Our proposed curriculum strategy significantly improves all three model performances across both datasets.} we\change{Unpaired one-tailed T-test is also conducted to check the significance of improvement for (i) SD over i.i.d, (ii) P-CD over SD, and (iii) P-CD over i.i.d on both datasets. Based on p-values (Table~\ref{tab:pvalue}), an improvement from employing P-CD over i.i.d is found to be very significant (p $\textless 0.05$) on both datasets in terms of DSC and IoU. Improvement from employing P-CD over SD is also found to be significant (p $\textless 0.1$) on the BUS dataset in terms of DSC.} Similar prediction trends are also observed in qualitative visualization in Fig~\ref{fig:qualitative_analysis}. A slight enhancement in boundary prediction is noticed in the visualization. The performance is also compared with the top-3 participants in the challenge~\cite{allan20202018} and our method outperforms with a remarkable margin (Fig.~\ref{fig:ComparisionandRobustness}(a)). \change{Interestingly, Top1 performing method of the challenge is also utilized DeepLabV3+, which yields 2-3\% higher mIoU by integrating our proposed P-CD. It is clear evidence that our technique is flexible and can be integrated with any state-of-the-art network to improve performance.}

\subsection{Robustness Test}

\begin{figure*}[!ht]
    \centering
    \includegraphics[width=0.90\linewidth]{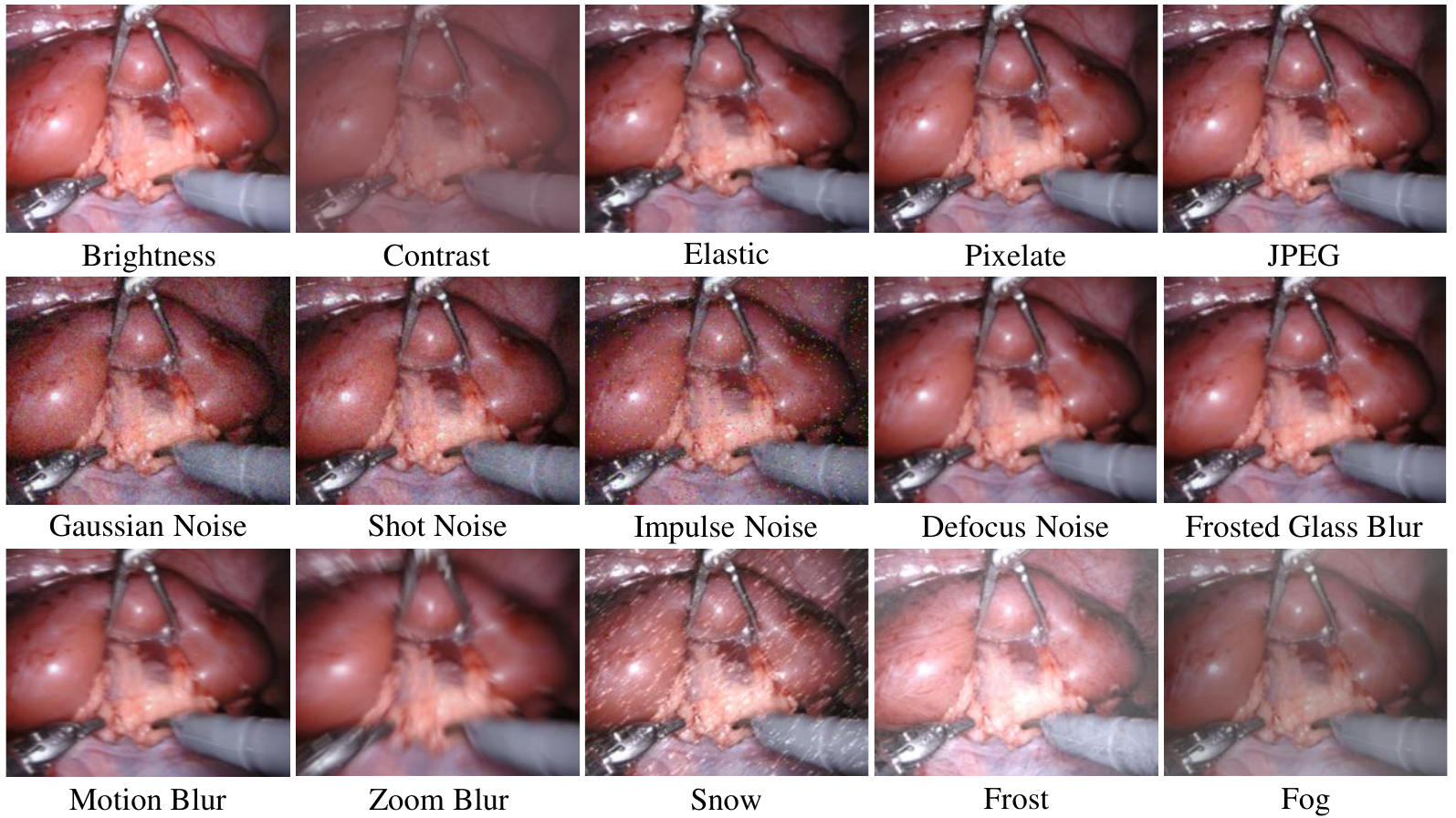}
    \caption{A surgical scene perturbed in 15 types with severity level set to $3$ to test the model robustness.}
    \label{fig:robustness_Dataset}
\end{figure*}

Models trained on curriculum are argued to be robust to noises~\cite{gong2016curriculum,wu2020curricula}. The robustness of our P-CD trained student models to common perturbations is quantified against models trained using i.i.d, KD technique is quantified by introducing corruptions and perturbations of varied severity by following~\cite{hendrycks2019benchmarking}. 
\change{There are 15 different perturbation techniques with 5 severity levels.  For example, noises such as Gaussian, shot, and impulse; blurring techniques such as defocus, glass, motion, zoom; weather corruptions such as snow, frost, and fog; and digital corruptions such as brightness, contrast, elastic, pixel, and JPEG. The severity levels are controlled by increasing the perturbation scale for each technique. For example, to add Gaussian noise with severity levels 1 to 5, the standard deviation of the Gaussian noise increase as $[0.04, 0.06, 0.08, 0.09, 0.10]$, respectively. The implementation of the perturbation techniques and severity levels are adopted from the reference work~\cite{hendrycks2019benchmarking} and official repository~\footnote{https://github.com/hendrycks/robustness}.} \changetwo{When segmenting a surgical field, the model must be robust to bleeding/tissue occlusion and occlusion from smoke. When the instrument or tissue is occluded/partially occluded by other instruments or bleeding or dead tissue or saline, the presence of other instrument features, translucent fluid, or dead tissues could significantly change the feature representation of the boundaries and result in boundary uncertainty. As the robotic surgery dataset lacks the annotation (corruption/severity of corruption) to quantify the model's robustness against them, we relate robustness against (a) weather \{fog\} to smoke occlusion, (b) noise \{gaussian\} and blurring \{defocus, glass, motion, zoom\} to saline release,  and  (C) digital \{pixel\} and noise \{gaussian\} to instrument occlusion.}

\subsection{Ablation Study}

\subsubsection{W and W\/O Boundary Uncertainty}
\begin{figure}[!h]
    \centering
    \includegraphics[width=1.0\linewidth]{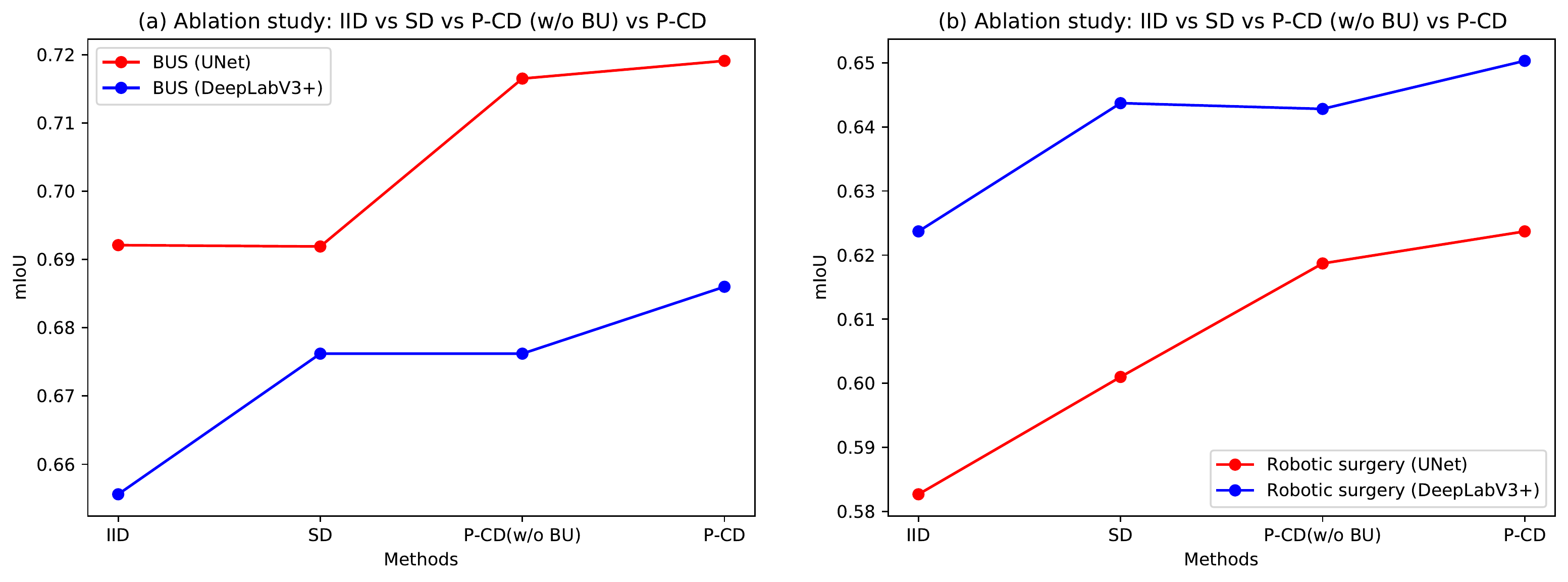}
    \caption{\change{Ablation studies: Performance improvements of UNet and DeepLabV3+ on (a) the BUS dataset and (b) the Robotic surgery dataset.}}
    \label{fig:Ablationstudy}
\end{figure}

Table~\ref{tab:main_table} presents the significant performance improvement with our P-CD. We also investigate the impact of model uncertainty and boundary uncertainty separately in \change{Fig.~\ref{fig:Ablationstudy}(a) \& (b)}. Although there is a clear prediction enhancement on curriculum learning with only model uncertainty, boundary uncertainty provides an additional benefit by difficulty measurement in curriculum strategy. This observation is further validated in Table~\ref{tab:pcd&BU}, where, both LinkNet~\cite{chaurasia2017linknet} and DeepLabV3+~\cite{chen2018encoder} trained using P-CD outperforms LinkNet~\cite{chaurasia2017linknet} and DeepLabV3+~\cite{chen2018encoder} trained using P-CD w/o BU.

\begin{table}[!h]
\centering
\caption{Performance comparison of models trained using P-CD w \& w/o BU for Robotic surgery dataset.}
\label{tab:pcd&BU}
\scalebox{0.9}{
\begin{tabular}{c|c|c|c|c}
\toprule
Model                       &             & Dice   & IoU    & Precision \\ \midrule
\multirow{2}{*}{LinkNet~\cite{chaurasia2017linknet}}    & P-CD w/o BU & 0.6432 & 0.6058 & 0.7161    \\ \cline{2-5} 
                            & P-CD        & \textbf{0.6715} & \textbf{0.6332 }& \textbf{0.7707}    \\ \midrule
\multirow{2}{*}{DeepLabV3+~\cite{chen2018encoder}} & P-CD w/o BU & 0.6822 & 0.6428 & \textbf{0.7902 }   \\ \cline{2-5} 
                            & P-CD        & \textbf{0.6929} & \textbf{0.6503} & 0.7803    \\ \midrule
\multirow{2}{*}{\change{UNet~\cite{ronneberger2015unet}}}       & \change{P-CD w/o BU} & \change{0.6594}      & \change{0.6187}      & \textbf{\change{0.7545}}         \\ \cline{2-5} 
                            & \change{P-CD}        & \textbf{\change{0.6644}} & \textbf{\change{0.6237}} & \change{0.7535}    \\ 
\bottomrule
\end{tabular}}
\end{table}

% \begin{table}[!h]
% \centering
% \caption{Performance comparison of model trained using P-CD w \& w/o BU for Robotic surgery dataset.}
% \label{tab:pcd&BU}
% \scalebox{0.95}{
% \begin{tabular}{cc|ccc}
% \toprule
% \multicolumn{2}{c|}{\multirow{2}{*}{\textbf{Model}}}                          & \multicolumn{3}{c}{\textbf{Robotic surgery}}                                                \\ \cline{3-5} 
% \multicolumn{2}{c|}{}                                                         & \multicolumn{1}{c|}{\textbf{Dice}} & \multicolumn{1}{c|}{\textbf{IoU}} & \textbf{Precision} \\
% \midrule
% \multicolumn{1}{c|}{\multirow{2}{*}{\textbf{LinkNet~\cite{chaurasia2017linknet}}}} & \textbf{P-CD w/o BU} & \multicolumn{1}{c|}{0.6432}        & \multicolumn{1}{c|}{0.6058}       & 0.7161             \\ \cline{2-5} 
% \multicolumn{1}{c|}{}                                  & \textbf{P-CD}        & \multicolumn{1}{c|}{\textbf{0.6715}}        & \multicolumn{1}{c|}{\textbf{0.6332}}       & \textbf{0.7707}             \\ 
% \midrule
% \multicolumn{1}{c|}{\multirow{2}{*}{\textbf{DeepLabV3+~\cite{chen2018encoder}}}} & \textbf{P-CD w/o BU} & \multicolumn{1}{c|}{0.6822}        & \multicolumn{1}{c|}{0.6428}       & \textbf{0.7902}             \\ \cline{2-5} 
% \multicolumn{1}{c|}{}                                  & \textbf{P-CD}        & \multicolumn{1}{c|}{\textbf{0.6929}}        & \multicolumn{1}{c|}{\textbf{0.6503}}       & 0.7803             \\
% \bottomrule
% \end{tabular}}
% \end{table}

\subsubsection{$\beta$ vs $\gamma$ Study}
\begin{figure}[!h]
    \centering
    \includegraphics[width=1.0\linewidth]{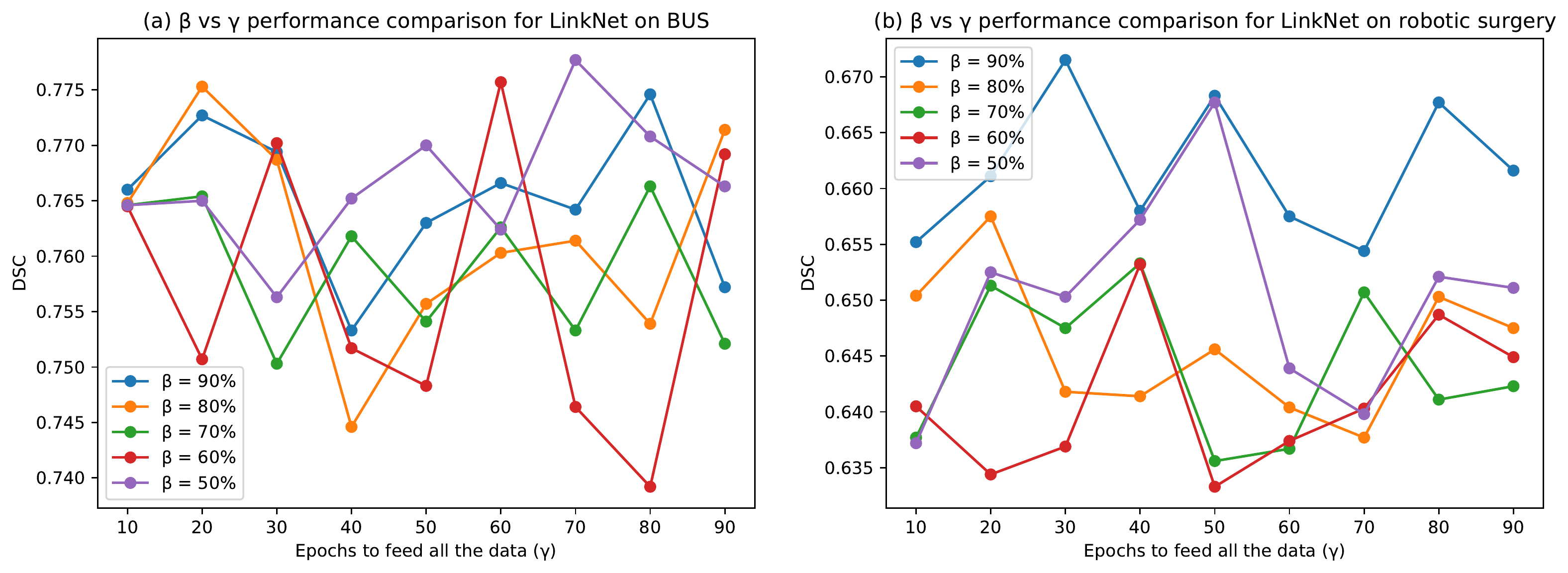}
    \caption{\change{$\beta$ vs $\gamma$ Study: $\beta$ and $\gamma$ tuning for LinkNet model on (a) the BUS dataset and (b) the Robotic surgery dataset.}}
    \label{fig:beta_vs_gamma}
\end{figure}

Initial $\beta$ and $\gamma$ values determine the initial $\mu$ value and dictate its increment during the training regime. As $\mu$ is used to pace the model learning, the effect on the model’s performance from $\beta$ and $\gamma$ is significant. We study and report the performance of the student \change{models} based on various $\beta$ vs $\gamma$ combinations on the BUS \change{and the Robotic surgery datasets (Fig.~\ref{fig:beta_vs_gamma}(a) \& (b)). While high performance is observed with $beta =90\%$ on the Robotic surgery dataset, it is not universal, as high performance on the BUS dataset is observed with $beta =50\%$.} Therefore, the best $\beta$ vs $\gamma$ could change for each dataset.

\section{Discussion and Conclusion} \label{Conclusion}
We present paced-curriculum distillation (P-CD), a model-agnostic curriculum-based distillation technique for semantic segmentation in medical imaging. We design a curriculum strategy by combining predicted uncertainty from a teacher model and label boundary uncertainty to measure pixel difficulty levels. Incorporating boundary uncertainty with prediction uncertainty firms the complex boundary pixels might assign wrong confidence in the less robust teacher model. We empirically show the superior performance of our method with two medical image segmentation datasets and extensive ablation studies. We also investigate the robustness of our method by corrupting validation data with 15 various types of image corruption and perturbation techniques. The results suggest that P-CD improves the performance and obtains better generalization and robustness over the dataset shift. \changetwo{In addition to overall performance improvement, the results from Table~\ref{tab:main_table} and  Fig.~\ref{fig:qualitative_analysis} demonstrate that P-CD is also able to predict better object boundaries, which is very crucial in clinical applications.} We observe that curriculum learning requires accurate hyper-parameters for pacing function, which leads to extensive tuning sometimes. However, the level of performance improvement suppresses this limitation. As future work, multi-raters disagreement~\cite{Wei_2021_WACV} can be integrated with P-CD to obtain a more accurate pixel difficulty score for robust curriculum distillation.

\section*{Declarations}

\subsection{Code availability}
This work is implemented using the PyTorch framework and the codes are available at: \href{https://github.com/mobarakol/P-CD}{github.com/mobarakol/P-CD}.

\subsection{Funding}
This work was supported by the Shun Hing Institute of Advanced Engineering (SHIAE project BME-p1-21) at the Chinese University of Hong Kong (CUHK), Hong Kong Research Grants Council (RGC) Collaborative Research Fund (CRF C4026-21GF and CRF C4063-18G), (GRS)\#3110167 and Shenzhen-Hong Kong-Macau Technology Research Programme (Type C 202108233000303).

\subsection{Ethics approval}
All procedures performed in studies involving human
participants were in accordance with the ethical standards of the institutional and/or national research committee.

\subsection{Consent to participate}
This article does not contain patient data.

\subsection{Conflict of interest} 
The authors declare that they have no conflict of interest.

\bibliography{sn-bibliography}% common bib file

\end{document}